\documentclass[times,twocolumn,final]{elsarticle}

\usepackage{cag}
\usepackage{framed,multirow}

\usepackage{caption}
\usepackage{subcaption}

\usepackage{amssymb}
\usepackage{latexsym}

\usepackage{url}
\usepackage{xcolor}
\definecolor{newcolor}{rgb}{.8,.349,.1}

\usepackage{hyperref}

\usepackage[switch,pagewise]{lineno} 
\usepackage{sg-ibs-macros} 

\journal{Computers \& Graphics}

\begin{document}

\verso{Preprint Submitted for review}

\begin{frontmatter}

\title{CloudWalker: Random walks for 3D point cloud shape analysis}%

\author[1]{Adi \snm{Mesika}\corref{cor1}}
\cortext[cor1]{Corresponding author: 
  E-mail: sadim@campus.technion.ac.il;}
    
\author[1,2]{Yizhak \snm{Ben-Shabat}\fnref{fn1}}
\fntext[fn1]{E-mail: sitzikbs@technion.ac.il}  
\author[1]{Ayellet \snm{Tal}\fnref{fn2}}
\fntext[fn2]{E-mail: ayellet@ee.technion.ac.il}  

\address[1]{Technion - Israel Institute of Technology, Haifa, Israel}
\address[2]{Australian National University, Canberra, Austrlaia}


\begin{abstract}
Point clouds are gaining prominence as a method for representing 3D shapes, but their irregular structure poses a challenge for deep learning methods. 
In this paper we propose CloudWalker, a novel method for learning 3D shapes using random walks. Previous works attempt to adapt Convolutional Neural Networks (CNNs) or impose a grid or mesh structure to 3D point clouds. This work presents a different approach for representing and learning the shape from a given point set. The key idea is to impose structure on the point set by multiple random walks through the cloud for exploring different regions of the 3D object. Then we learn a per-point and per-walk representation and aggregate multiple walk predictions at inference. Our approach achieves state-of-the-art results for two 3D shape analysis tasks: classification and retrieval. 
\end{abstract}

\begin{keyword}
\KWD point cloud, neural networks, 3d shape classification, 3d shape retrieval
\end{keyword}

\end{frontmatter}


\begin{figure*}[tb]
    \includegraphics[width=0.98\linewidth]{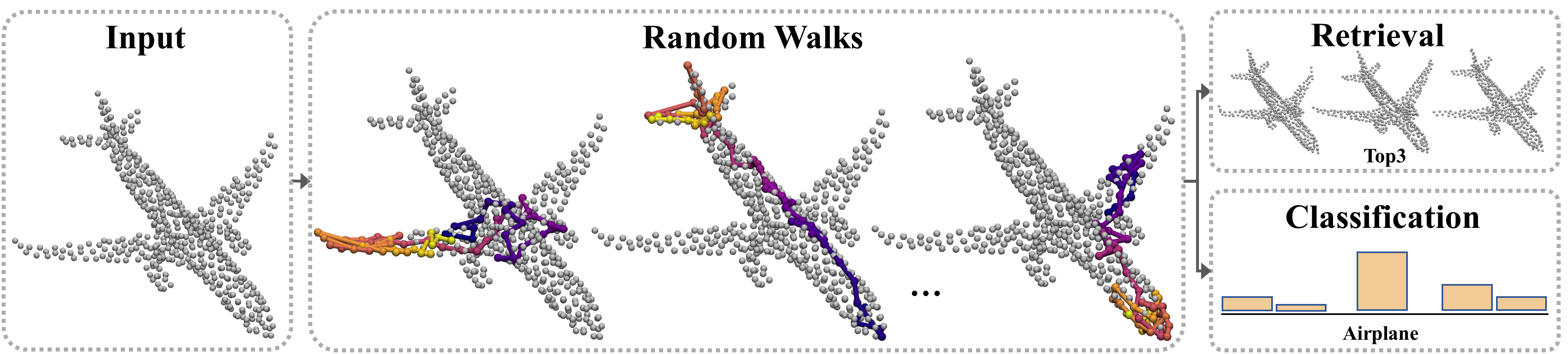}
    \caption{{\bf CloudWalker's pipeline.} A set of random walks is generated for a given point cloud. 
    Each walk explores the object (walk points are color-coded according to the walk sequence, from blue to yellow). We use CloudWalker for two tasks: 3D object retrieval and classification and achieve SOTA results. 
    }
    \label{fig:pipeline}
\end{figure*}

\section{Introduction}

\label{sec:intro}

Three dimensional data acquired by sensors provides rich geometric information. 
Various formats are used to represent this data, including meshes, volumetric grids, multi-view images, implicit functions and point clouds \cite{guo2020deep, forsyth2011computer, marschner2018fundamentals}.
Our focus is on point cloud representation, which is a sampling of a continuous surface. It is often the rawest form of data obtained by modern 3D scanners. 

Point clouds manage to preserve the original geometric information in 3D space without the need of discretization. 
However, there are challenges associated with this representation, particularly in the context of deep learning approaches.
These include the lack of connectivity information, structure and order, the varied number of points, and corruptions such as noise, incompleteness, and density variations. 
An additional difficulty is the lack of large-scale datasets.

Several approaches have been recently proposed for point cloud deep learning. 
Point-wise {\em Multilayer perceptron (MLP)} methods  explore global/local structures to enhance feature learning and overcome the permutation variance (order) using a symmetric function~\cite{qi2017pointnet,qi2017pointnet++,zaheer2017deep} . 
Others propose convolutional based methods, which can be divided into continuous~\cite{thomas2019kpconv, xu2018spidercnn, liu2019relation,wu2019pointconv,liu2019densepoint} and discrete~\cite{li2018pointcnn, rao2019spherical}.
These methods overcome some of the structural challenges using local connectivity information.
Some works consider each point as a vertex of the graph, by adding edges that connect points, and apply graph networks~\cite{simonovsky2017dynamic}, while others construct a  hierarchical data structures ({\em e.g.}, Octree and KD-tree)~\cite{lei2019octree,klokov2017escape}.
While all of the above works handle the challenges of lack of structure and order, none has shown to be as effective as CNNs are for images.
Therefore, the challenge of learning a representation for point clouds remains an active field of research.

Our work, illustrated in \figref{fig:pipeline}, takes a different avenue.
We propose to represent a point cloud by random walks through the cloud, which ”explore” the shape.
Intuitively, such a walk moves from point to point, through the cloud's valleys and ridges, and on its way discovers the geometry of the underlying surface.
A walk is  essentially a sequence of points.
In a way, this sequence imposes some structure, which is essential for deep learning.
A point cloud is represented by multiple walks, which partially cover the underlying surface.
Due to their randomness, some walks provide a more global view of the object, whereas others are local and focus on specific parts.

Recently, random walks were used for a variety of 3D model representations. MeshWalker \cite{lahav2020meshwalker} proposed to use random walks on meshes, however, random walks on point clouds is significantly different and poses several challenges, addressed in our work. First, point clouds lack the connectivity information that is inherently available in meshes, therefore there is no notion of local environment for each point. We solve this issue using a KD-tree data structure that uses spatial proximity to propose local points as candidates for walk generation.  Moreover, since a mesh vertex differs from a point (due to connectivity information), our per-point representation is somewhat different from MeshWalker's vertex representation, by adding parameters, layers and additional possible neighbour connections. Lastly, the walks for each shape are aggregated differently than the walks for each mesh, as we are less certain regarding consecutive points than in the case of vertices. The fact that our local environment is not unequivocal has an impact on performance, which we demonstrate in the ablation study.

Xiagn \emph{et al.} ~\cite{xiang2021walk} proposed to use guided walks on point clouds. They generated curves based on a given set of rules and heuristics and then learn the shape representation, for that purpose, they get the entire point set as input. Our approach is significantly different and simpler since
we only get part of the data (walk per shape) and utilize the power of randomness. In spite of the simplicity of our approach, we show that we can produce good results. 

We show how to overcome major challenges in deep learning based point cloud processing:
First, we address the generation of  walks despite the lack of connectivity information and show how to do that efficiently.
Furthermore, unlike some previous algorithms \cite{qi2017pointnet, qi2017pointnet++, xu2018spidercnn}, our approach does not require normal vectors as inputs to improve performance. This is an important property since normal estimation is particularly inaccurate for the available datasets that contain synthetic CAD models or noisy real-world point clouds without normal \cite{guerrero2018pcpnet, ben2020deepfit, lenssen2020deep}. Moreover, we learn a per walk and per point representation and propose an effective cross-walk aggregation module to produce a global shape representation. 

Our approach is general in the sense that it can be used for various shape analysis tasks.
We demonstrate its performance on two tasks: 3D object classification and 3D object retrieval.
Our results are compared against the reported SOTA results for commonly-used  datasets 
{\em 3D-Future}~\cite{fu20203dfuture},  
{\em ScanObjectNN}~\cite{uy-scanobjectnn-iccv19}, and
{\em ModelNet40}~\cite{wu20153d}.

\noindent This paper makes two main contributions:
\begin{itemize}
\item  We propose a novel representation for point clouds using random walks. The walks impose order and overcome the lack of connectivity and structure. 
\item  We present an end-to-end learning framework that realizes this representation. We show that it works well even when the dataset has few unique objects and lacks normals. We achieves SOTA results for 3D shape classification and retrieval.
    
\end{itemize}

\section{Related work}

A variety of representations of 3D point cloud have been proposed in the context of deep learning. The main challenge is how to re-organize the set description such that it could be processed within deep learning frameworks. Hereafter we briefly review the main representations; see \cite{guo2020deep} for a recent thorough survey.

Generally, 3D data for shape classification methods is represented using one of the following representations: (a) Multi-view, (b) Volumetric grid, (c) Mesh, (d) Point clouds. Each representation requires a different
approach for modifying the data to the form required by
deep learning methods. 

\textbf{Multi-view approaches.} Multi-view-based methods project the 3D data into 2D images and extract view-wise features, and then fuse these features
for accurate shape classification. MVCNN \cite{su2015multi} is a pioneering work, which essentially max-pools multi-view features into a global descriptor. Some information is lost in the projection process, but using multiple projections partially compensates. In addition, View-GCN \cite{wei2020view} represented multiple views of a 3D shape by a view-graph. The view-graph representation enables to design Graph Convolutional Neural Network (GCN) to aggregate multi-view features by investigating relations of views. 

\textbf{Volumetric grid approaches.} An alternative approach to transforming irregular point clouds to regular representations is 3D voxelization \cite{maturana2015voxnet,qi2016volumetric,wu20153d}, followed by 3D Convolution Neural Network (CNN). If applied naively, this strategy can incur massive computation and memory costs due to the cubic growth of the number of voxels as a function of resolution. Since most voxels are empty, the solution is to take advantage of sparsity. For example, OctNet \cite{riegler2017octnet} uses unbalanced octrees with hierarchical partitions. An approach based on sparse convolutions that only evaluate occupied voxels would significantly reduce computation and memory requirements.

\textbf{Triangle mesh approaches.} A mesh is represented as a set of vertices, edges and faces. To handle meshes directly, novel convolutions vertex neighborhoods have been defined \cite{feng2019meshnet, verma2018feastnet}. Other works parameterize the mesh in 2D \cite{boscaini2016learning,ezuz2017gwcnn, sinha2016deep}. In MeshCNN \cite{hanocka2019meshcnn}, a unique idea of using the edges of the mesh to perform pooling and convolution, is introduced. Our work is inspired by MeshWalker, which takes a different approach of representing a mesh by a set of random walks over its vertices, along the edges \cite{lahav2020meshwalker}.

\textbf{Point Cloud approaches.} The point cloud representation is challenging because it is both unstructured and point-wise unordered. To overcome these challenges, PointNet \cite{qi2017pointnet, qi2017pointnet++} utilizes permutation-invariant operators such as pointwise MLPs and pooling layers to aggregate features across a set. A number of approaches connect the point set into a graph and conduct message passing on this graph. DGCNN \cite{wang2019dynamic} performs graph convolutions on kNN graphs. KCNet \cite{shen2018mining} utilizes kernel correlation and graph pooling. DeepGCNs \cite{li2019deepgcns} explore the advantages of depth in graph convolutional networks for 3D scene understanding.
GDANet \cite{xu2021learning} use an attention network and introduces a Geometry-Disentangle Module to dynamically disentangle point clouds into the contour and flat part of 3D objects. A voxel-based point cloud transformer, PVTNet \cite{zhang2021pvt}, introduced a Sparse Window Attention module to alleviate the problem of expensive computations
In addition, many other schemes have also been proposed.
PCEDNet \cite{himeur2021pcednet} focused on classification of edges in point clouds. They proposed a new parameterization for each point in a set containing millions of points.
RCNet \cite{wu2019point} utilizes standard RNN to construct a permutation-invariant network for 3D point cloud processing. Point2Sequences \cite{liu2019point2sequence} is another RNN-based model that captures correlations between different areas in local regions of point clouds. Here we proposed the CloudWalker which is directly linked to the point cloud representation and can be used as input for a neural network.

\textbf{Traditional random walks methods.} Several methods utilized random walks for shape analysis in both the computer vision and graphics communities. Prior to the deep learning era they were frequently used for image clustering and segmentation \cite{meilua2001random, gorelick2006shape, grady2006random, grady2004multi, grady2005multilabel, lai2008fast}. They were also used for generating super-pixels \cite{shen2014lazy} and for visual tracking \cite{li2015visual}. For a comprehensive survey of random walks applications see \cite{xia2019random}.

\section{CloudWalker}

\begin{figure*}[t]
    \includegraphics[width=0.95\linewidth]{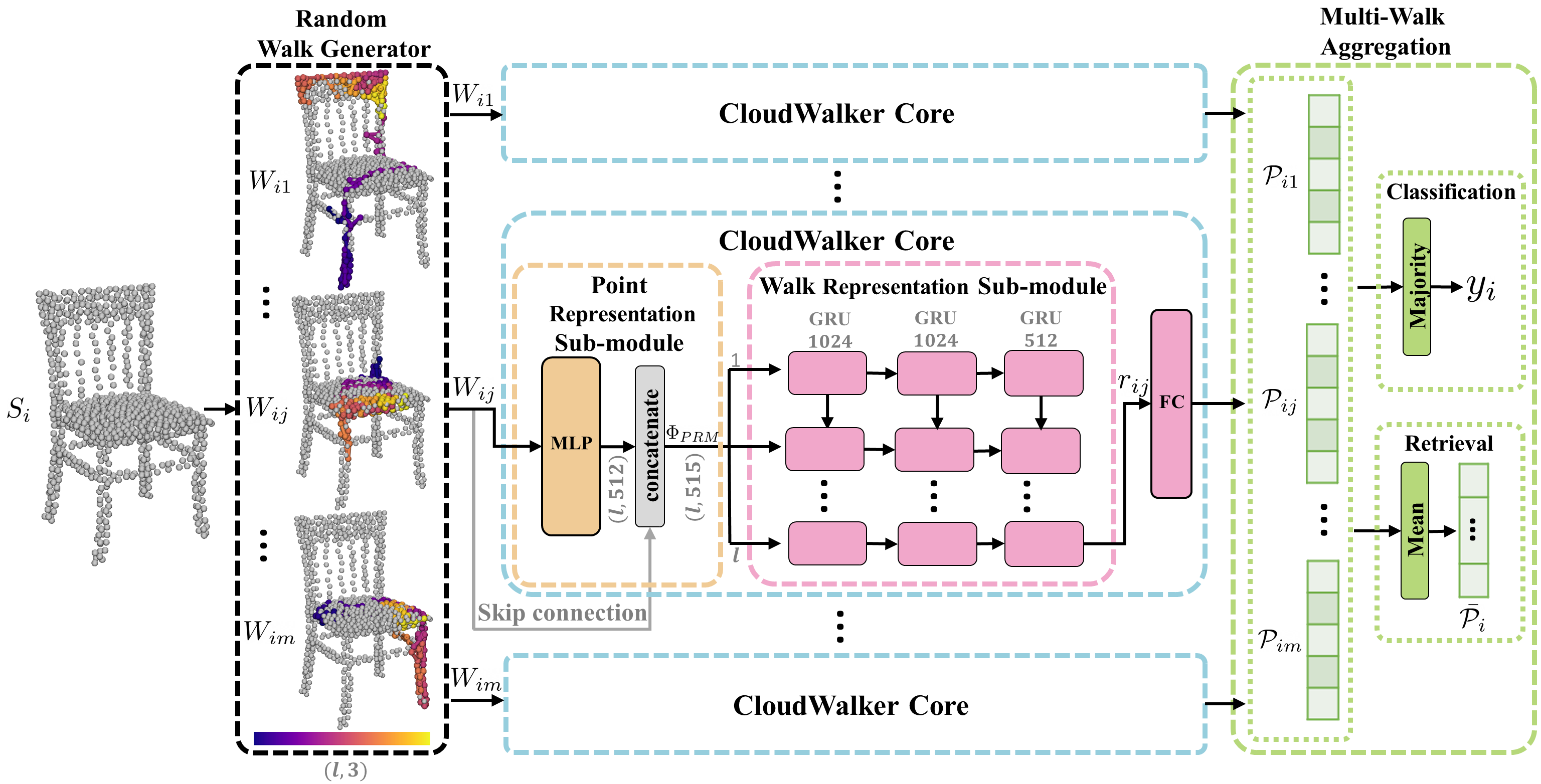}
    \caption{{\textbf{CloudWalker's pipline.} 
    For an input point set $S_{i}$, the {\em Random Walks Generator} generates several walks $W_{ij}$. 
    Our model learns a representation for each walk by learning a shared, per-point representation, which are then aggregated along the walk by the {\em Walk Representation} sub-module.
    This sub-module consists of three RNN layers and produces a walk representation $r_{ij}$. 
    The {\em Multi-Walk Aggregation} module is used at inference time to combine several walks into either the final class prediction for classification or an average representation for retrieval. 
    }}
    \label{fig:model}
\end{figure*}

Given a point cloud, our goal is to learn a representation that will capture both global and local geometric information about the underlying shape, for  analysis tasks. Previous works showed that learning 3D object's representation is improved by combining both local and global features~\cite{qi2017pointnet++}.
As discussed above, point clouds are challenging for deep learning approaches, due to the lack of spatial structure, which prohibits the use of spatial filters on a 3D lattice.
Furthermore, point clouds come in various sizes.

We propose to address these challenges by using multiple random walks through the clouds. Each walk wanders through the object and "explores" it on the way. Jointly, multiple walks provide both local shape information, as well as global shape information; the very same region can be visited multiple times from different directions. This satisfies the above desired properties since the walk length is independent w.r.t. the number of points in the cloud (up to full coverage) and the points have a sequential order, that covers both local and global information. 

Our proposed model consists of three modules, as illustrated in \figref{fig:model}: 
The {\em Random Walk Generator Module} generates multiple random walks per point set. 
The {\em CloudWalker Core} module extracts a single walk representation.
It consists of two sub-modules: the {\em Point Representation Sub-module (PRM)} learns a shared, per-point representation, and  
the {\em Walk Representation sub-module (WRM)} combines the information along the walk and learns the relations between the points. 
Finally, the {\em Multi-walk Aggregation Module} produces the final shape representation, by aggregating multiple walk representations. 
This representation can then be used for shape analysis tasks.
Hereafter we elaborate on each of these modules.

\noindent
\textbf{Random walk generator.} 
The concept of randomness is very powerful~\cite{motwani1995randomized, chazelle2001discrepancy, sawhney2020monte}.
This module generates multiple random walks for a given point set. 
In our setup, thanks to randomness some walks explore a single region of the point set and learn its fine details alongside some broader context,
while other walks cover more regions and provide the model with additional information, not seen by other walks. 

A walk is a sequence of points, where each point is associated with basic local geometric information.
Should the underlying surface of the point cloud existed, it would be possible to utilize the adjacency information to determine, for each point, the next point on the walk.
However, since connectivity information is unavailable, 
we propose a fast and simple approach to generate random walks. 

For an input point set $S_i$ we generate the walk $W_{ij}$ of length $l$ as follows.
First, the walk's origin point $p_o \in S_i$ is randomly selected from the set.
Then, points are iteratively added to the walk by selecting a random point from the set of $k$ nearest neighbors of the last point in the sequence (excluding neighbors that were added at an earlier stage). 
In the rare case where all $k$ nearest neighbors were already added to the walk, a new random un-visited point is chosen and the walk generation proceeds as before. Choosing the closest neighbor might seem like a straight-forward solution, however, this imposes a strong constraint on the generation process and reduces the randomness and the ability to visit sparser regions.

Since a point cloud may contain many points, an efficient nearest neighbor search is required to get a sense of the environment and propose point candidates to generate the walk. In contrast to meshes, where this information is inherently available, point clouds require this critical step to generate the walks.
We propose to construct a KDtree \cite{de1997computational} as a preprocessing step.
Briefly, a KDtree is a hierarchical space partitioning data structure, in which every node represents an axis-aligned hyper-rectangle and contains the set of points in it.
This is a very efficient data structure for nearest neighbor queries. Using  the nearest neighbors to construct the walk makes it possible to move between underlying connected components along the walk, which works to our advantage since it allows our model to learn such crossings.

\noindent
\textbf{CloudWalker Core.} 
Given an input walk, CloudWalker core is the main learned part of our pipeline. It is designed to learn an informative representation for a single walk. it consists of two sub-modules, \textit{Point Representation} sub-module and \textit{Walk Representation} sub-module. Details for each sub-module are provided below.

\noindent
\textbf{Point representation sub-module. } 
Given an input walk $W_{ij}$ (the $j^{th}$ walk for point cloud $S_i$), each point $p_t$ along the walk is embedded onto a high-dimensional feature space $\phi_{PRM}(p_t)$.
We convert each point to a $512$-dimension vector using a three layered Multi-Layer Perceptron (MLP) that shares weights across points. This module is different from MeshWalker's vertex representation module by an additional MLP layer. The additional parameters give the network more capacity to compensate for the lack of connectivity information and learn a distinctive per-point representation. Then, we perform instance normalization and ReLU nonlinear activation.
We preserve the original point context by concatenating the input coordinate to the per-point representation and then feeding it into the next sub-module (WRM); see \figref{fig:model}.
The module's output is a sequence of vectors per walk, $\Phi_{PRM}(W_{ij})$.

\noindent
\textbf{Walk representation sub-module. } 
The input to this stage is a sequence of high-dimensional feature vectors, a single vector for each point in the walk.
The output of this module is the walk representation $r_{ij} = \Phi_{WRM}(\Phi_{PRM}(W_{ij}))$ in the form of a single feature vector for the walk.

Similarly to~\cite{lahav2020meshwalker}, we use three {\em Gated Recurrent Units (GRU)}~\cite{cho2014learning}.
Briefly, a GRU is able to "remember" and accumulate knowledge in a sequence. In our setup, each GRU layer receives all the embedded features from all points of the walk until the current point. 
It outputs the last hidden state, which captures an abstract representation of the input sequence. In a sense, this is the descriptor that describes the entire walk.
This architecture allows walk information to persist. 
It is especially powerful since the local geometric detail as well as global geometric context are critical for describing the shape.

Finally, the walk representation $r_{ij}$ is fed into a fully connected (FC) classifier to produce the prediction vector $\widetilde{r_{ij}}$. The number of elements in $\widetilde{r_{ij}}$ is equal to the number of classes.

\noindent
\textbf{Multi-walk aggregation module.} 
This module gets as input multiple walk predictions $ \widetilde{r_{ij}}, j \in [1,m]$ of a single shape $S_i$. 
To compute the probability vector we apply a softmax: $\mathcal{P}_{ij} = softmax(\widetilde{r_{ij}})$. 

Then, we apply a symmetric aggregation function over all of the input walks. 
For the symmetric function we chose the majority vote:
Thus, for a given $S_i$, the output prediction $y_i$ is: 
\begin{equation}
y_i = Majority(\mathcal{C}_{i1},\ldots,\mathcal{C}_{ij},\ldots,\mathcal{C}_{im}),
\end{equation}
where $\mathcal{C}_{ij}=argmax(\mathcal{P}_{ij})$ is the class prediction.

We have experimented with alternative options for symmetric function, including $max$ and $mean$, and found the majority vote to work best.
This is attributed to its robustness to cases where the network is overconfident in a small subset of walks. A quantitative ablation study is presented in \secref{sec:ablation}.

\noindent {\bf Implementation details.}
At inference, we use $m=48$ walks per object $S_i$, 
the length of the walk is $l=800$,
and $k=20$ nearest neighbors. 
In training, we use Adam optimizer with cyclic learning rate; the initial and the maximum learning rates are set to $10^{-6}$ and $5 \cdot10^{-4}$ respectively. The cycle size is $20K$ iterations and we train for a total of 100K iterations. We record the number of parameters and time using TensorFlow2.4 on a single NVIDIA GeForce RTX 3090. The total number of network parameters is 13.5M and training process takes 100 epochs that last 97 seconds each on average (on ModelNet40 \cite{wu20153d}). The average inference time is 0.22 second per shape. The distances between predictions and ground truth labels were minimized through cross entropy loss. Note that sometimes ({\em e.g.} ModelNet40) the shape scale is available and may be informative. In this case we add the shape's bounding box diagonal length to the feature vector and feed it to the classifier for a minor improvement. Code will be released upon acceptance. 

\section{Applications}
\label{sec:applications}
We evaluate the performance of our model on two main tasks: 3D shape classification and 3D object retrieval.

\subsection{3D shape Classification}
\label{subsec:classification}

Given a 3D point set, the goal is to classify it into one of pre-defined classes. 
Our model  outputs a classification prediction probability vector for each walk. 
At inference, a majority prediction of the walks' prediction vectors  is  calculated to produce the final result.

\noindent
\textbf{Evaluation measures. }
For each dataset we report both {\em instance accuracy (IA)} and {\em class accuracy (CA)}. 
Instance accuracy is defined as the percentage of the correctly-classified objects, 
\begin{equation}
    IA=\frac{\#correctly\_classified}{\#test\_instances}
\end{equation}
Class accuracy is the mean of per-class instance accuracy, \begin{equation}
    CA=\frac{1}{C}\sum_{i\in{C}}{IA(i)},
\end{equation}
where $C$ is the number of classes.

\begin{figure}[tb]
    \begin{subfigure}{1\linewidth} \centering
        \includegraphics[width=0.95\textwidth]{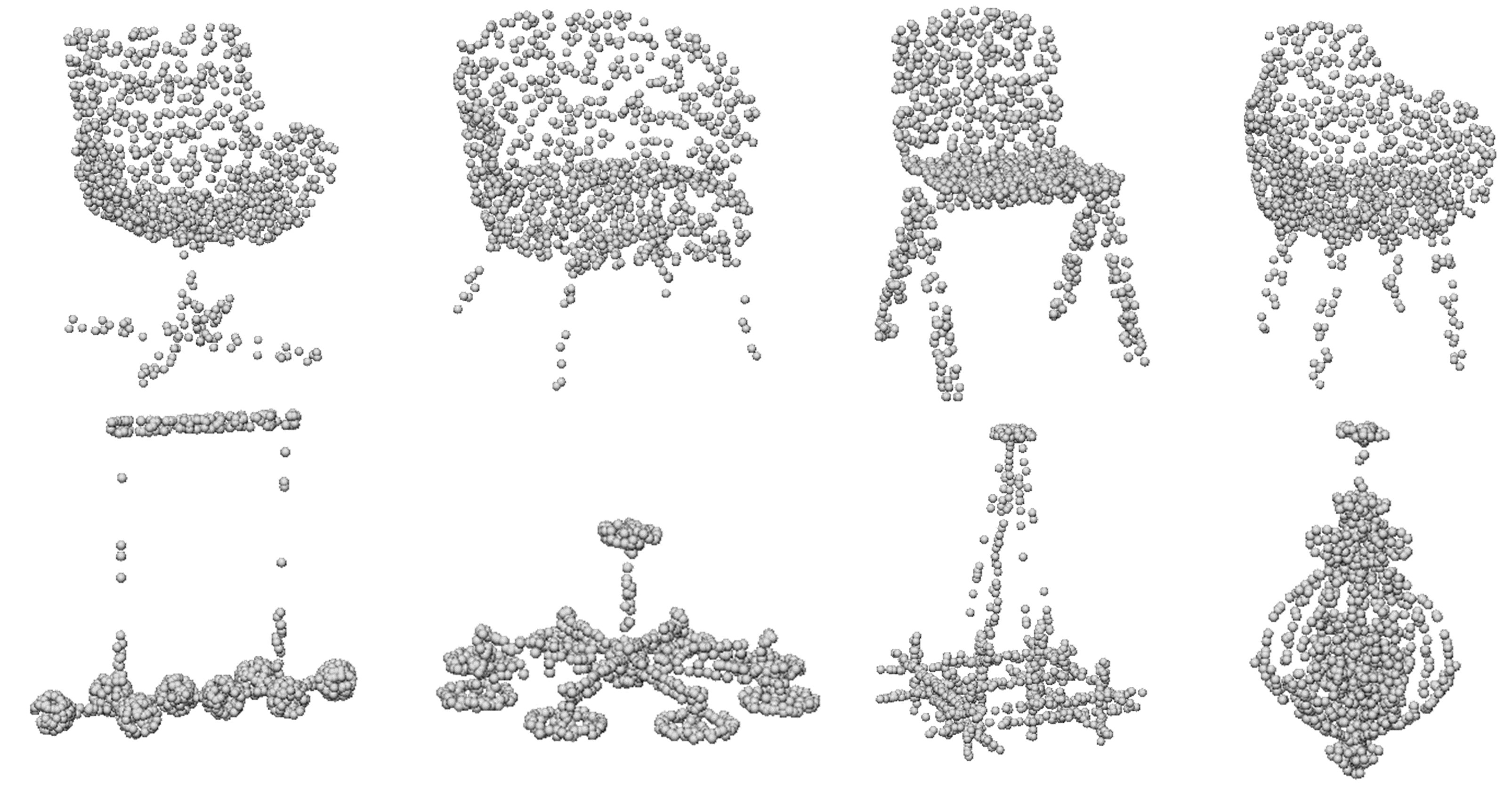}
        \caption{Examples from 3D-FUTURE dataset \cite{fu20203dfuture}}
        \label{fig:3dfuture-examples}
    \end{subfigure}
  \begin{subfigure}{1\linewidth} \centering
        \includegraphics[width=0.95\textwidth]{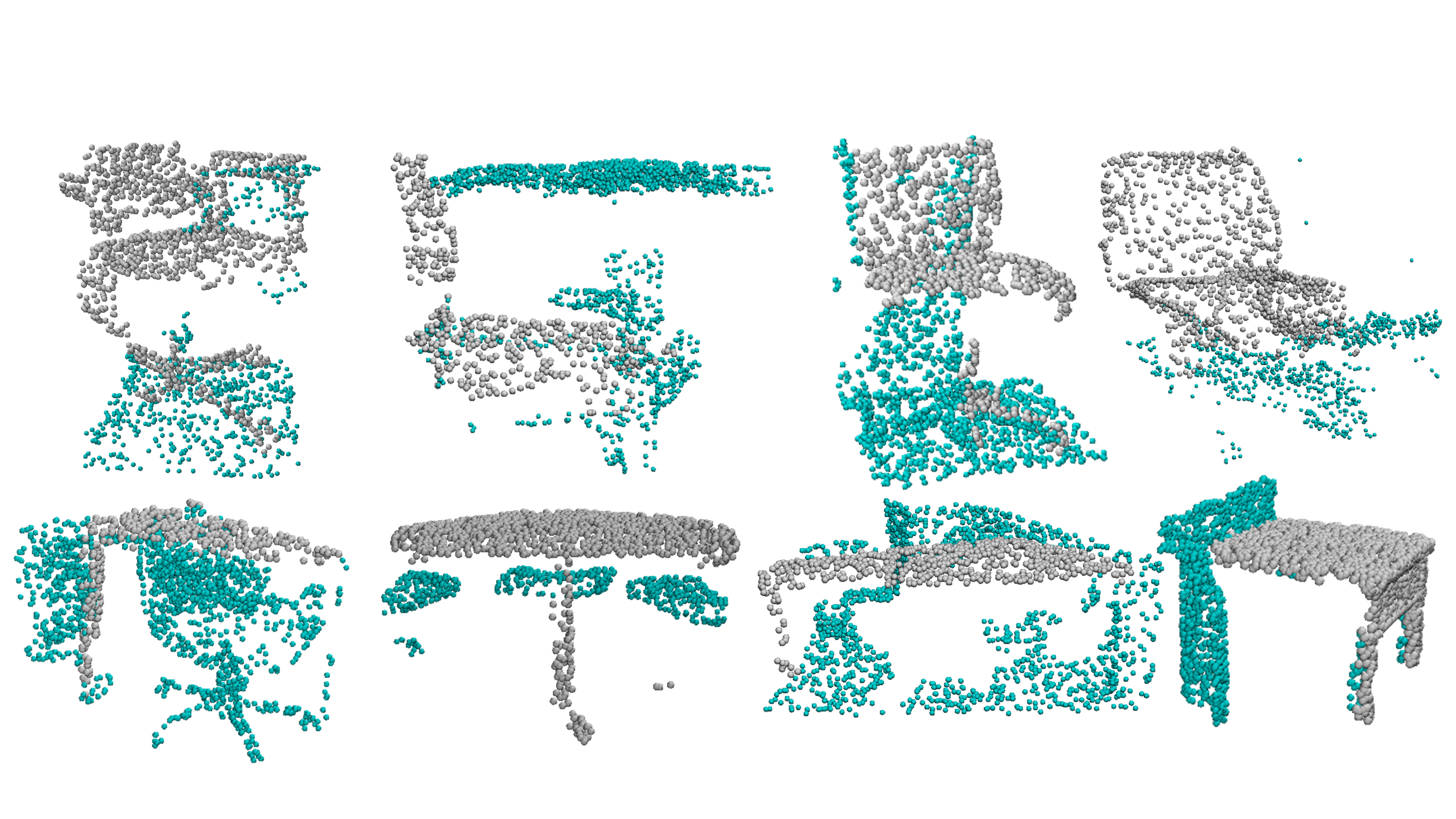}
        \caption{Examples from ScanObjectNN dataset \cite{uy-scanobjectnn-iccv19}}
        \label{fig:scanobjectnn-examples}
    \end{subfigure}
    \begin{subfigure}{1\linewidth} \centering
        \includegraphics[width=0.95\textwidth]{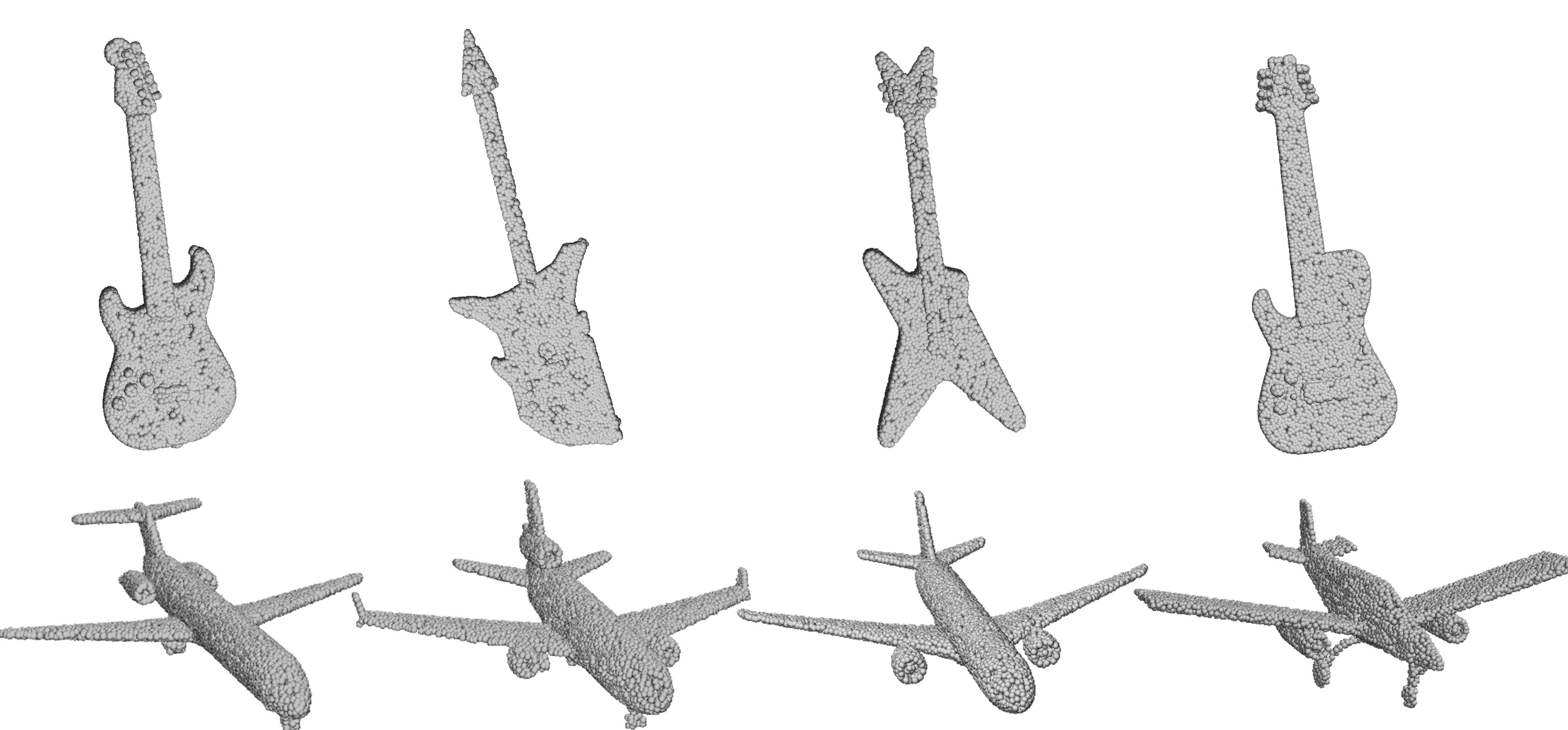}
        \caption{Examples from ModelNet40 dataset \cite{wu20153d}}
        \label{fig:modelnet40-examples}
    \end{subfigure}
  \hfill
  \caption{{\bf Datasets snapshot. } Datasets vary in density, noise and available classes. Presented here are typical objects from \subref{fig:3dfuture-examples} 3D-FUTURE which contains uniformly sampled industrial CAD models (chair and lamp classes), \subref{fig:scanobjectnn-examples} ScanObjectNN which contains real-world  objects, with their background in cyan (chair and table classes), and \subref{fig:modelnet40-examples} ModelNet40 which contains CAD models (guitar and airplane classes).}
  \label{fig:datasets-examples}
\end{figure}

\noindent\textbf{Datasets.}
We present our results on three commonly-used datasets, where each contains different types of objects, different numbers of classes, a different number of objects per class, and different challenges.
\figref{fig:datasets-examples} presents some examples from the datasets.

\noindent
{\em 3D-FUTURE}~\cite{fu20203dfuture} is a recent dataset that contains $9,992$ industrial CAD models of furniture . It consists of $7$ super-categories, with $1$-$12$ sub-categories each, for a total of $34$ categories. The train/test split is $6,699$/$3,293$. This dataset is challenging both due to the objects it contains and its hierarchical structure, as objects in the same category but different sub-categories may resemble each other, requiring fine-grained information in the shape representation. 

\noindent
{\em ScanObjectNN}~\cite{uy-scanobjectnn-iccv19} is a real-world dataset that contains 
$2902$ unique object scans from $15$ classes. The objects are corrupted in various ways to create a set of ${\sim}15k$ objects that are used for training and testing. We follow the official data split provided in \cite{uy-scanobjectnn-iccv19}.
This dataset is highly challenging due to the data corruptions found in scans which include: background points, object partiality, and different transformation variants (translation, rotation, and scale). We performed our experiment on its most challenging variant, {\em PB T50 RS}, with and without background points. 

\noindent
{\em ModelNet40}~\cite{wu20153d} contains CAD models. There are $12,311$ meshes divided into $40$ classes, with $9,843$/$2,468$ shapes for training/testing. Corresponding point clouds are generated by uniformly sampling, following the sampling protocol of~\cite{qi2017pointnet++}.

\noindent
\textbf{Quantitative results.}
The results for 3D-FUTURE \cite{fu20203dfuture} are presented in Table \ref{tab:3DFUTURE-short_table}. 
It shows that our method outperforms not only previous point-based, but also multi-view methods, reported for this dataset. 
We followed the same training protocol for point-based methods as specified by~ \cite{fu20203dfuture} and trained our model with $1K$ points per object. 
In particular, we uniformly sampled $1024$ points for each shape.
Our results outperform the others in both IA and CA evaluation measures.

\tabref{tab:3DFUTUREResults} presents quantitative result for classification on 3D-FUTURE for each category. It shows that our method outperforms both, point-based and multi-view methods reported for this dataset. Our model leads on average and in $18/32$ categories. 
It shows how challenging this dataset is because of its similar sub-categories. 
We note that while \cite{qi2017pointnet++} uses normals as additional information, we do not.
The mere vertices locations suffice for our method.
This is especially important in cases where normals are unavailable, as normals are not outputs of most scanners and normal estimation is possible, but is not a trivial problem \cite{guerrero2018pcpnet,ben2020deepfit,lenssen2020deep}.
After all, normal estimation is "almost" surface reconstruction, and the reconstruction of a surface is exactly what we wish to avoid.

\begin{table}[tb] 
  \caption{{\bf Classification results on \textit{3D-FUTURE} dataset.} Our results outperform those reported in \cite{fu20203dfuture} using both multi-view and point clouds as input. We note PointNet++ uses normals while we only use points. (* Our reproduction, the orginal paper did not report on this dataset)}
\centering
  \begin{tabular}{||c|c|c|c||}
    \hline
    Method & Input & IA (\%) & CA (\%)\\
    \hline
    MVCNN \cite{su2015multi} & Multi-views & $69.2$ & $65.4$ \\
    CurveNet* \cite{xiang2021walk} & Point cloud & ${69.6}$ & ${65.1}$ \\
    PointNet++ \cite{qi2017pointnet++} & Point cloud & $\underline{69.9}$ & $\underline{66.0}$ \\
    \hline
    \hline
    Ours & Point cloud & $\textbf{70.6}$ & $\textbf{67.8}$ \\
  \hline
  \end{tabular}
  \label{tab:3DFUTURE-short_table}
\end{table}

\begin{table}[tb] \addtolength{\tabcolsep}{-4pt}  
\centering
  \caption{{\bf Classification results for each category of \textit{3D-FUTURE}}. 
  We note PointNet++ (PN++) uses normals and Multi-scale grouping (MSG) and we do not, while we use only the points. MVCNN uses 12 view.}
  \begin{tabular}{||c|c|c|c||} 
    \hline
    Category & Ours (\%) & PN++ (\%)& MVCNN (\%)\\
    \hline
    Children Cabinet  & $16.0$ & $\underline{32.1}$ & $\textbf{72.0}$\\
    Nightstand  & $\textbf{83.5}$ & $71.8$ & $\underline{75.0}$\\
    Bookcase  & $\underline{58.7}$ & $52.3$ & $\textbf{66.7}$\\
    Wardrobe   & $\textbf{85.0}$ & $\underline{82.0}$ & $56.7$\\
    Coffee Table  & $\textbf{83.7}$ & $\underline{82.6}$ & $67.9$\\ 
    Corner/Side Table  & $\underline{73.0}$ & $\textbf{74.7}$ & $64.5$\\
    Side Cabinet  & $\textbf{69.5}$ & $\underline{65.2}$ & $47.9$\\ 
    Wine Cabinet   & $44.3$ & $\textbf{67.1}$ & $\underline{62.9}$\\ 
    TV Stand   & $\textbf{80.5}$ & $\underline{73.6}$ & $73.5$\\
    Drawer Chest   & $44.8$ & $\underline{55.2}$ & $\textbf{67.5}$\\ 
    Shelf  & $\underline{48.4}$ & $\underline{48.4}$ & $\textbf{51.9}$\\ 
    Round End Table  & $\textbf{93.7}$ & $\underline{75.0}$ & $52.2$\\
    King-size Bed   & $\underline{89.6}$ & $\textbf{91.2}$ & $78.6$\\ 
    Bunk Bed   & $\textbf{88.8}$ & $\underline{77.8}$ & $57.1$\\
    Bed Frame   & $\textbf{95.4}$ & $\underline{93.8}$ & $\underline{93.8}$\\
    Single bed & $\textbf{68.9}$ & $\textbf{68.9}$ & $\underline{67.7}$\\
    Kids Bed  & $00.0$ & $\textbf{14.3}$ & $\underline{12.5}$\\
    Dining Chair & $\textbf{76.8}$ & $\underline{63.9}$ & $50.5$\\
    Lounge/Office Chair  & $54.0$ & $\textbf{60.5}$ & $\underline{60.3}$\\
    Classic Chinese Chair  & $\textbf{62.5}$ & $\textbf{62.5}$ & $\underline{57.1}$\\
    Barstool  & $\textbf{88.8}$ & $\underline{66.7}$ & $32.0$\\
    Dressing Table  & $\textbf{77.3}$ & $68.2$ & $\underline{73.7}$\\ 
    Dining Table  & $\underline{76.0}$ & $61.1$ & $\textbf{84.8}$\\
    Desk  & $\underline{30.9}$ & $20.4$ & $\textbf{54.0}$\\ 
    Three-seat Sofa  & $\textbf{83.1}$ & $\underline{82.6}$ & $71.7$\\ 
    armchair  & $\textbf{74.9}$ & $68.0$ & $\underline{72.5}$\\ 
    Loveseat Sofa  & $45.5$ & $\textbf{64.5}$ & $\underline{62.9}$\\ 
    L-shaped Sofa   & $\textbf{92.9}$ & $\underline{85.9}$ & $83.3$\\
    Lazy Sofa  & $42.8$ & $\underline{50.0}$ & $\textbf{66.7}$\\
    Stool & $70.9$ & $\underline{75.8}$ & $\textbf{91.9}$ \\
    Pendant Lamp & $\textbf{92.8}$ & $\underline{90.9}$ & $89.8$\\
    Ceiling Lamp  & $\textbf{72.0}$ & $\underline{70.7}$ & $63.0$\\
    \hline
    \hline
    Average Class Accuracy & $\textbf{67.8}$ & $\underline{66.0}$ & $65.2$\\
    Instance Accuracy & $\textbf{70.6}$ & $\underline{69.9}$ & $69.2$\\
    \hline
  \end{tabular}
  \label{tab:3DFUTUREResults}
\end{table}


For ScanObjectNN, Table \ref{tab:scanobjectnnResults} presents the results with and without background points of real-world scans.
It shows that the overall accuracy is better compared to state of the art methods~\cite{uy-scanobjectnn-iccv19}. 
When trained and tested without the background points, 
our average class accuracy $78.5\%$ is higher than all reported results while maintaining competitiveness on the overall accuracy.

\begin{table}[tb] 
  \caption{{\bf Classification results on \textit{ScanObjectNN} \cite{uy-scanobjectnn-iccv19}.}
  We use the difficult variant of the dataset, PB T50 RS, with (w/) and without (w/o) background (BG) points. 
  We achieve SOTA results on the variant without the background;
   ($CA$ results were not reported for objects without background).
  we achieve CA SOTA results on the variant with the background and competitive IA results. The second best is underlined. (*concurrent work) }
\centering
  \begin{tabular}{||c|cc|cc||}
    \hline
     \multirow{2}{*}{Methods}& \multicolumn{2}{c|}{w/o BG (\%)} & \multicolumn{2}{c||}{w/ BG (\%)} \\
      & IA & CA & IA & CA\\
    \hline
    PointNet \cite{qi2017pointnet} & $74.4$ & - & $68.2$ & $63.4$ \\
    PointNet++ \cite{qi2017pointnet++} & $80.2$ & - & $77.9$ &	$75.4$ \\
    3DmFV \cite{ben20183dmfv} & $69.8$ & - & $63.0$ & $58.1$ \\
    SpiderCNN \cite{xu2018spidercnn} & $76.9$ & - & $73.7$ & $69.8$ \\
    SimpleView \cite{goyal2020revisiting} & -  & - & ${79.5}$ & - \\
    DGCNN \cite{wang2019dynamic} & $\underline{81.5}$ & - & $78.1$ & $73.6$ \\ 
    PointCNN \cite{li2018pointcnn} & $80.8$ & - & $78.5$ & $75.1$ \\
    DRNet \cite{qiu2021dense} & - & - & $80.3$ & $\underline{78.0}$ \\
    GBNet \cite{qiu2021geometric} & -  & - & $\underline{80.5}$ & $77.8$ \\
    DeltaConv* \cite{wiersma2021deltaconv} & - & - & $\textbf{84.7}$ & - \\
    \hline
    \hline
    BGA-PN++ \cite{uy-scanobjectnn-iccv19}& -  & -  & $80.2$ & $77.5$ \\
    BGA-DGCNN \cite{uy-scanobjectnn-iccv19}& -  & -  & $79.7$ & $75.7$ \\
    \hline
    \hline
    Ours & $\textbf{82.2}$ & $79.5$ & $80.3$ & $\textbf{78.5}$ \\
    \hline
  \end{tabular}
  \label{tab:scanobjectnnResults}
\end{table}

\begin{figure}[tb]
 \centering
    \includegraphics[width=0.85\linewidth]{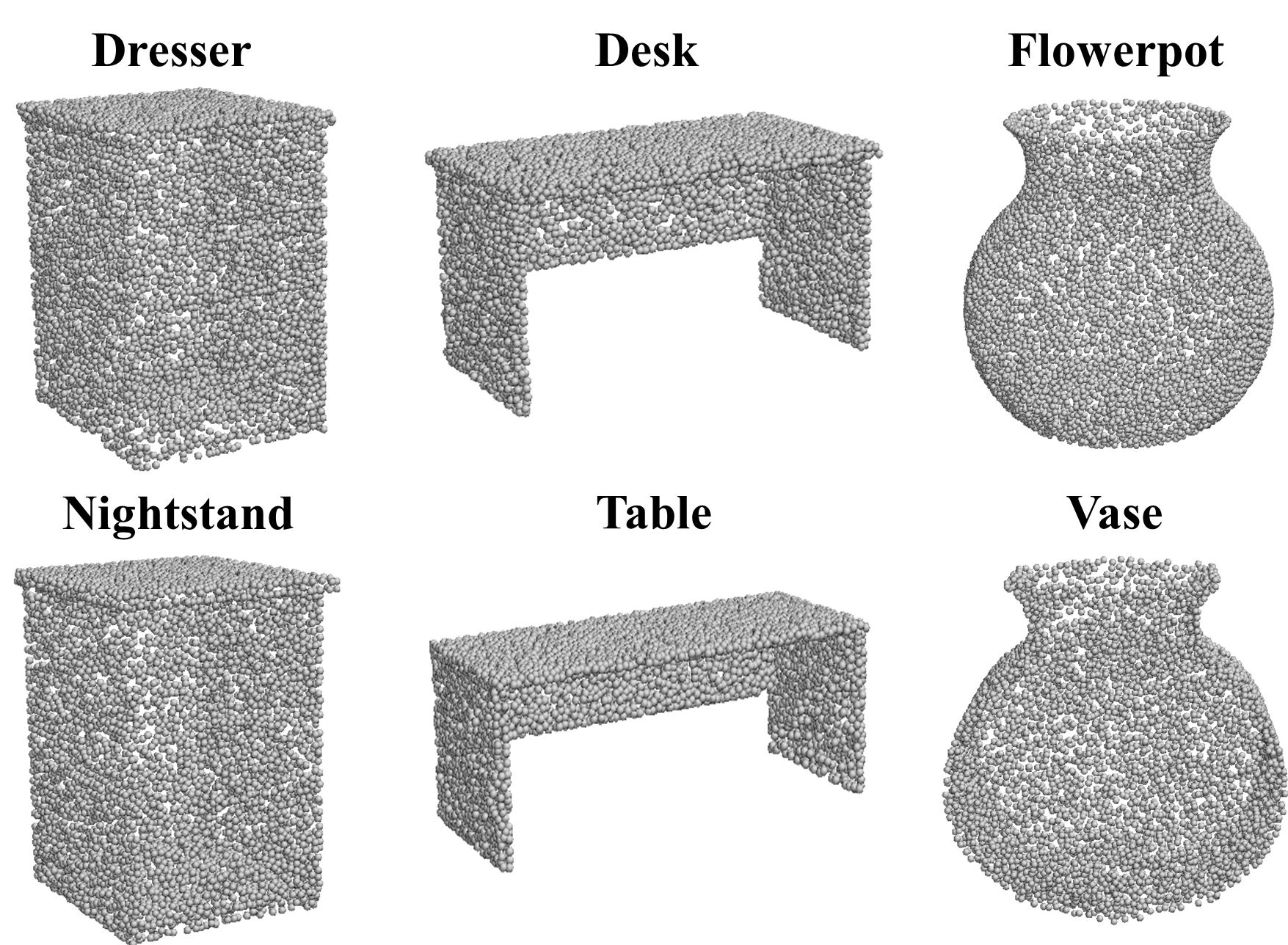}
  \caption{{\bf ModelNet40 \cite{wu20153d} difficult examples.} Objects from different classes have similar geometric properties.}
  \label{fig:Modelnet40-SimilarObjectDiffClasses}
\end{figure}

Table \ref{tab:Modelnet40Results} reports the results on the ModelNet40 \cite{wu20153d} dataset.
For this dataset we achieved competitive results.
ModelNet40 is known to be a difficult and saturated dataset, partially because of cross-labeled  classes (desk/table, plant/flower-pot/vase, night-stand/dresser), as illustrated in \figref{fig:Modelnet40-SimilarObjectDiffClasses}. 
 
 We report that when we choose the second best prediction for the mis-classified objects in Modelnet40 dataset, $97\%$ of the test instances are correctly classified. Similarly, for the 3DFuture dataset we get 89\% accuracy (${\sim}19\%$ improvement). This finding further shows that some categories have fine-grained differences that are difficult to discriminate.
 
\begin{table}[tb] 
  \caption{{\bf Classification results on \textit{ModelNet40} \cite{wu20153d}}
  For this dataset we show competitive results.}
\centering
  \begin{tabular}{||c|c|c||}
    \hline
    Method & IA (\%)& CA (\%)\\
    \hline
    PointNet \cite{qi2017pointnet} & $89.2$ & $86$ \\
    PointNet++ \cite{qi2017pointnet++} & $90.7$ & - \\
    PointCNN \cite{li2018pointcnn} & $92.2$ & $88.1$ \\
    SpiderCNN \cite{xu2018spidercnn} & $92.4$ & - \\
    KPConv \cite{thomas2019kpconv} & $92.9$ & - \\
    SimpleView \cite{goyal2020revisiting} & $93.6$ & $90.5$ \\
    PAConv \cite{xu2021paconv} & $93.6$ & - \\
    PointTrans. \cite{zhao2020point} & $93.7$ & $\underline{90.6}$ \\
    CurveNet \cite{xiang2021walk} & $93.8$ & - \\
    GBNet \cite{qiu2021geometric} & $93.8$ & $\textbf{91}$ \\
    PVT \cite{zhang2021pvt} & $\underline{94.0}$ & - \\
    RPNet \cite{ran2021learning} & $\textbf{94.1}$ & - \\
    \hline
    \hline
    Ours & $93.1$ & $90.1$ \\
    \hline
  \end{tabular}
  \label{tab:Modelnet40Results}
\end{table}

\subsection{Retrieval}
Given a query object, the goal is to retrieve objects from the dataset, ordered by their relevancy (similarity) to the query.
There are various ways to evaluate shape retrieval.
The most common evaluation is the {\em mean Average Precision (mAP)} over test queries:
\begin{equation}
    mAP=\frac{1}{Q}\sum_{i=1}^{Q} AP(S_i),
\end{equation} 
where $Q$ is the number of queries in the set. $AP$ is defined as $AP=\frac{1}{GTP}\sum_{k}^{N}{P@k}\times{rel@k},$ where $GTP$ is the number of ground truth positives, $N$ is the size of the ordered set, ${P@k}$ refers to the precision at $k$ and $rel@k$ is an indicator function which equals 1 if the object at rank k is from the same class as query $S_i$ and 0 otherwise. 
Note that significantly fewer methods report their performance on this task than for classification.

We present our results on commonly used datasets for shape retrieval task, ModelNet40  and ModelNet10 \cite{wu20153d}. 
We use the most common train/test splits: $9,843/2,468$ for ModelNet40 and $3,991/908$ for ModelNet10.
We used the output feature vector from our classification network by averaging all walks probability vectors for each object and get a global feature vector $\bar{\mathcal{P}_{i}}$; see \figref{fig:model}.

Table \ref{tab:ModelnetRetrievalResults} shows that our method achieves SOTA results on both ModelNet40 and ModelNet10 \cite{wu20153d}. 
It outperforms approaches that use a variety of 3D representations, including meshes, multi-views, and point clouds.

\begin{table}[tb] \setlength{\tabcolsep}{4pt}
  \caption{{\bf 3D Shape Retrieval results.}
  This table shows the $mAP$ on ModelNet40 \& ModelNet10~\cite{wu20153d}, sorted by the input type (Point Cloud, Multi-view, and mesh).
  Our CloudWalker outperforms other methods.
  }
\centering
  \begin{tabular}{||c|c|c|c||}
    \hline
    Method & input & ModelNet40 & ModelNet10 \\
    \hline
    GWCNN \cite{ezuz2017gwcnn} &Mesh& $59.0$ & $74.0$ \\
    MeshNet \cite{feng2019meshnet} &Mesh& $81.9$ & - \\
    \hline
    \hline
    MVCNN \cite{su2015multi} &MV& $79.5$ & - \\
    SeqViews \cite{han2018seqviews2seqlabels} &MV& $89.1$ & - \\
    \hline
    \hline
    PointNet \cite{qi2017pointnet} & PC& $70.5$ & -\\
    PointCNN \cite{li2018pointcnn} &PC& $83.8$ & -\\
    DGCNN \cite{wang2019dynamic} &PC& $85.3$ & -\\
    DensePoint \cite{liu2019densepoint} &PC& $88.5$ & $93.2$ \\
    \hline
    \hline
    Ours & PC& $\textbf{92.9}$ & $\textbf{93.7}$ \\
    \hline
  \end{tabular}
  \label{tab:ModelnetRetrievalResults}
\end{table}

\begin{figure}[tb] \centering
    \includegraphics[width=0.95\linewidth]{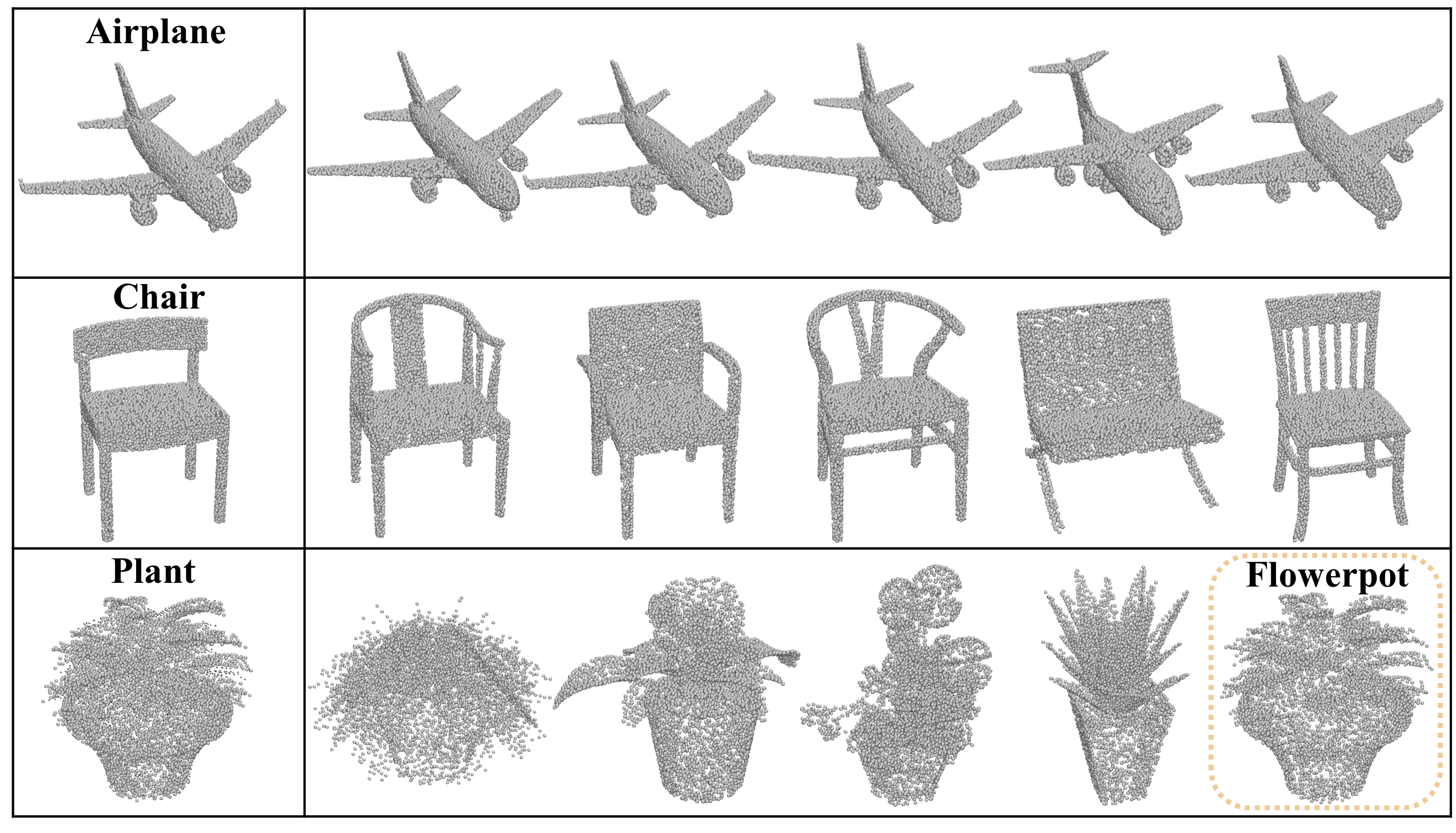}
  \caption{{\bf Qualitative results for Object Retrieval.} 
  The query test objects are on the left, where the top five retrieved objects are on the right.
  The falsely retrieved object is highlighted.}
  \label{fig:Modelnet40-retrieval_Qualitative_results}
\end{figure}

\figref{fig:Modelnet40-retrieval_Qualitative_results} presents qualitative results for object retrieval on ModelNet40 \cite{wu20153d} dataset. 
It shows the top five retrieved objects by our model. 
As can be seen, the retrieved objects are indeed similar and thus belong to the same class.
In the bottom row we show a particularly interesting case where the query object belongs to the \textit{plants} category, but the highlighted model belongs to a different category and thus considered to be an error. However, this is in fact the exact same object with different set of points. This example demonstrates some of the challenges in the dataset.
 
\section{Ablation study}
\label{sec:ablation}
\noindent
\textbf{Walk length.} In this experiment we explore the effects of walk length, {\em i.e.} the number of points in each walk, on the performance of classification. Table \ref{tab:ablation_study_walk_length} shows that a longer walk improves performance up to a certain length.
 This can be explained by the fact that longer walks are likely to have better coverage of the object, and thus better distinguish between different categories. However, when the walk is too long, the GRUs' ability to "remember" points in the sequences is reduced, which negatively impacts performance.
For example, on ScanObjectNN \cite{uy-scanobjectnn-iccv19}, we get improvement when using up to $40\%$ of the total number of vertices per walk. This quantity is dataset dependant, however we found $40\%$ to perform well across multiple datasets. 
\begin{table}[tb] 
  \caption{{\bf Walk length ablation.} 
  Performance improves as length increases, up to a certain length. Tested on ScanObjectNN~\cite{uy-scanobjectnn-iccv19}.}
\centering
  \begin{tabular}{||c|c|c||}
    \hline
    Walk length & IA (\%) & CA (\%) \\
    \hline
    $0.1V$ & $75.3$ & $72.1$\\
    $0.2V$ & $78.0$ & $75.2$\\
    $0.4V$ & $\textbf{82.2}$ & $\textbf{79.5}$\\
    $0.5V$ & $79.3$ & $76.4$\\ 
    \hline
  \end{tabular}
  \label{tab:ablation_study_walk_length}
\end{table}

\noindent
\textbf{Aggregation method.} 
In our model, we aggregate predictions from multiple walks on the shape.
Table \ref{tab:ablation_study_Aggregation_method} compares three  symmetric aggregation functions. 
While the results are similar, there is a slight advantage to majority voting. 

\begin{table}[tb] 
  \caption{{\bf Aggregation method ablation.} Majority voting is slightly better than other symmetric functions. Tested on ModelNet40~\cite{wu20153d}.}
\centering 
  \begin{tabular}{||c|c|c||}
    \hline
    Aggregation method  & IA (\%) & CA (\%)\\
    \hline
    Average & $92.8$ & $89.9$\\
    Max & $92.4$ & $89.2$\\
    Majority & $\textbf{93.1}$ & $\textbf{90.1}$\\
    \hline
  \end{tabular}
  \label{tab:ablation_study_Aggregation_method}
\end{table}

\noindent
\textbf{Number of  walks.} 
We study how the number of walks (used at inference) influences classification accuracy.
\tabref{tab:ablation_study_number_of_walks} shows that as the number of walks increases, the accuracy improves up to 48 walks where it saturates. Note that above 48 walks most points in the point cloud are visited at least once ($\sim$full coverage). Note that, even very few walks result in very good accuracy. This means that the number of walks may be a tuneable parameter that balances the trade-off between computational power and accuracy. In our method, we chose to use 48 walks. 
\begin{table}[tb] 
  \caption{{\bf Number of walks ablation.} The accuracy improves with the number of walks per shape. Tested on 3D-FUTURE \cite{fu20203dfuture}.
  }
\centering
  \begin{tabular}{||c|c|c|c||}
    \hline
    Number of walks & Coverage (\%) & IA (\%) & CA (\%)\\
    \hline
    $4$  & $76$ & $68.6$ & $65.5$\\
    $16$ & $82$ & $69.7$ & $67.1$\\
    $32$ & $92$ & $69.8$ & $67.5$\\
    $48$ & $99$ & $\textbf{70.6}$ & $\underline{67.8}$\\
    $64$ & $99$ & $\underline{70.5}$ & $67.5$\\
    $96$ & $100$ & $70.3$ & $\textbf{67.9}$\\
    \hline
  \end{tabular}
  \label{tab:ablation_study_number_of_walks}
\end{table}

\noindent
\textbf{Random walk generation method.}
Generating random walks can be performed in various ways. 
In the following we explore several options, in all of which the first point is randomly chosen and the next points are sequentially added from the last point's neighbors.
The options are:
(1) Random---randomly choosing among the point's unvisited neighbors with a uniform distribution.
(2) High variance---calculating the change in variance for each neighbor of the most recently added point and selecting the one that increases the variance the most. 
(3) Combined---combining both random and high variance methods by choosing a neighbor who increases the variance $30\%$ of the times and randomly otherwise. 
Note that the heuristic based on the walk variance is effective, but it is not sufficient since it reduces randomness.
The results are presented in Table \ref{tab:ablation_study_generation_method}. We found that the random strategy is the best, reinforcing our claim of the power of randomness.
\begin{table}[tb] 
  \caption{{\bf Walk generation ablation.} Random walk generation performs better than hand-crafted heuristics. Tested on ModelNet40~\cite{wu20153d}.
  }
\centering
  \begin{tabular}{||c|c|c||}
    \hline
    Generation method  & IA (\%) & CA (\%) \\
    \hline
    Random & $\textbf{93.1}$ & $\textbf{90.1}$ \\
    High variance & $90.2$ & $86.5$ \\
    Combined & $91.7$ & $87.9$ \\
    \hline
  \end{tabular}
  \label{tab:ablation_study_generation_method}
\end{table}

\noindent
\textbf{Stability.}
Cloudwalker relies on the power of randomness. To evaluate the consistency across different random seeds (stability), we evaluate our model 100 times, each time, we generate a different set of random walks. We found that the average overall accuracy for ModelNet40 dataset is $92.8\%$, with a standard deviation of $0.1\%$ and a maximum value of $93.1\%$. 

\begin{figure}[tb]
    \begin{tabular}{p{0.95\linewidth}}
    \includegraphics[width=0.95\linewidth]{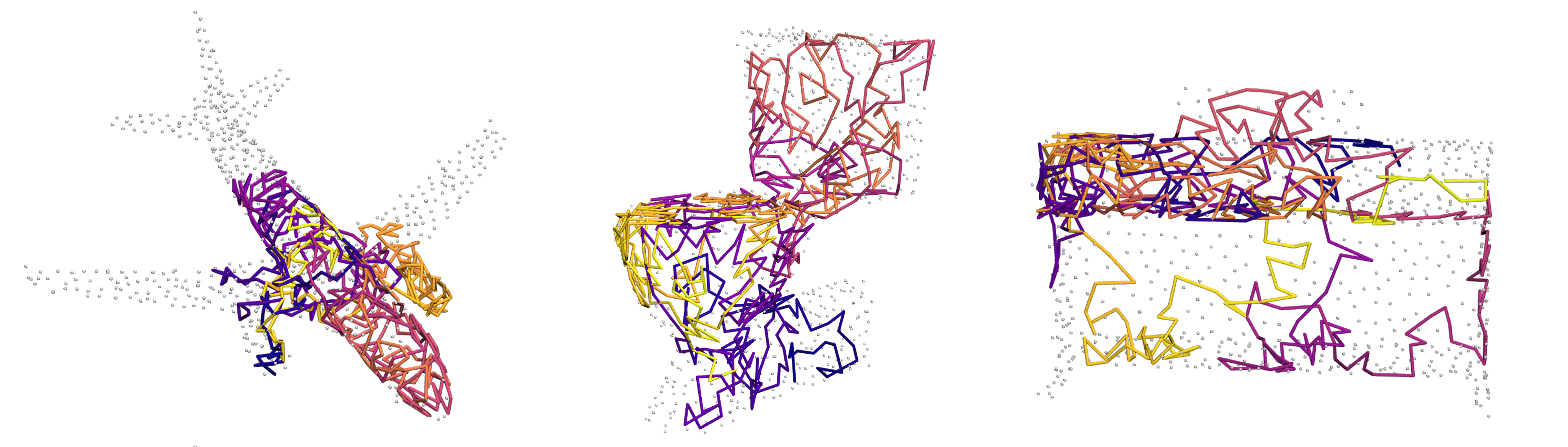}\\
    \includegraphics[width=0.95\linewidth]{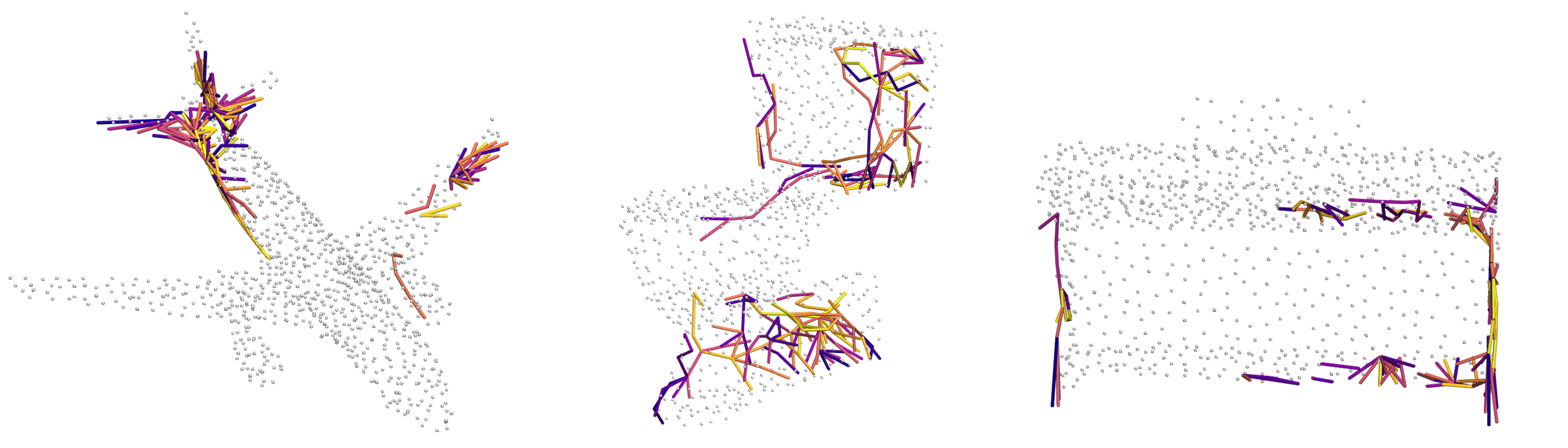}\\
    \end{tabular}
\caption{{\bf Walk comparison}. 
    CloudWalker and CurveNet walks on three different shapes. Top row: A single CloudWalker walk of length 500 points (color-coded according to the walk sequence, from blue to yellow), bottom row: A single CurveNet walk, composed of 100 curves of length 5 points (different color for each curve).}
  \label{fig:walkComparison}
\end{figure}

\noindent
\textbf{Comparison of random and guided walks.}
\figref{fig:walkComparison} compares our random walks to the guided walks generated by CurveNet~\cite{xiang2021walk} on the same object. It is evident that each method explores different parts of the model. Interestingly, although the walks are distinctly different, both methods manage to successfully classify the shape. Note that we restrict our walks from revisiting vertices, which is different from CurveNet, where there is no such hard restriction and curves may overlap in accordance with a crossover suppression strategy.

\begin{figure}[tb]
    \centering
    \begin{tabular}{c}
    \includegraphics[width=0.95\linewidth]{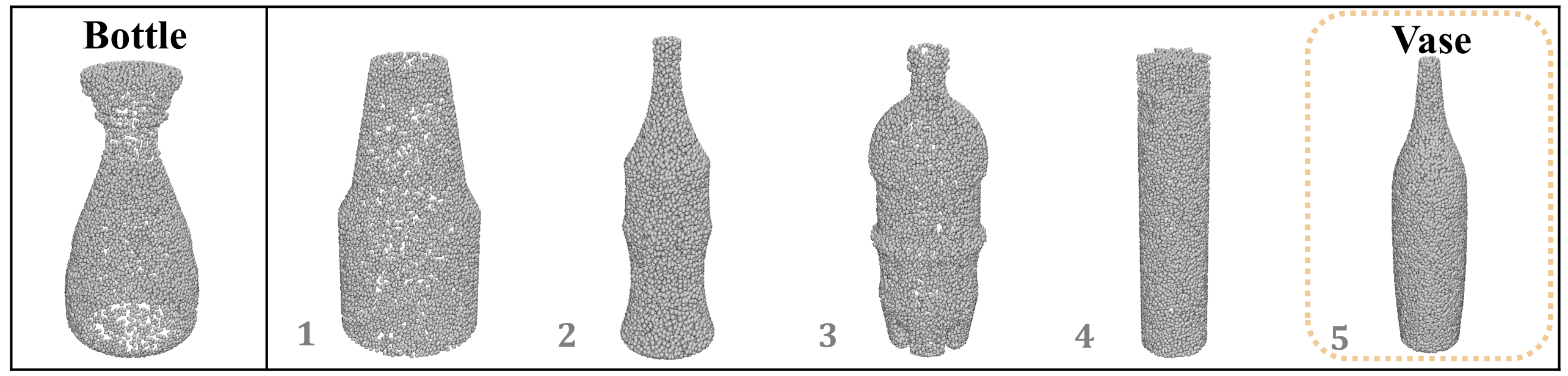}\\
    \small{(a) Same geometry with a different small detail (top cap).}\\
    \includegraphics[width=0.95\linewidth]{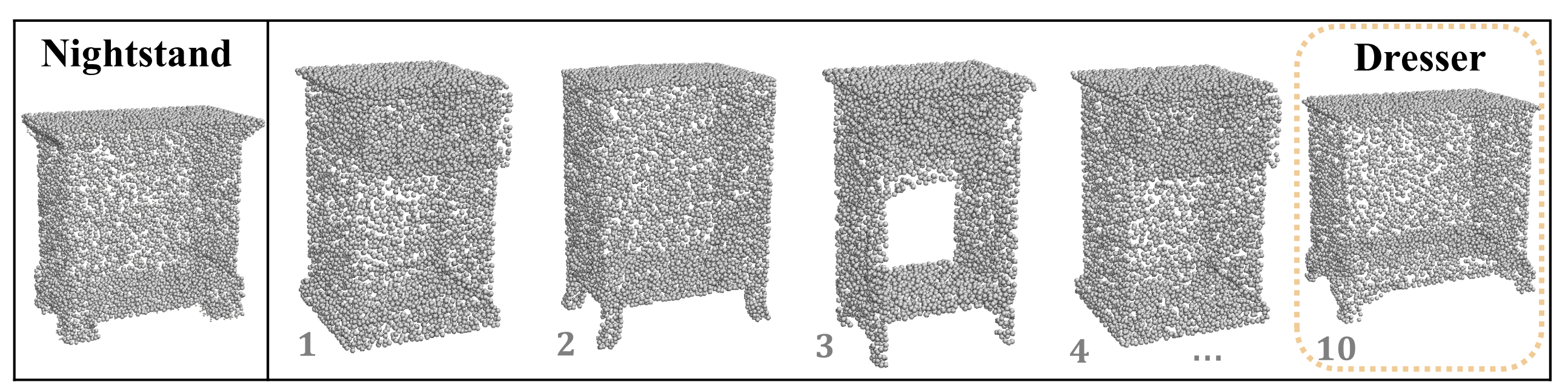}\\
    \small{(b) Geometrically similar objects are in different categories.}\\
    \end{tabular}
    \caption{{\bf Limitation}. 
    When objects are geometrically similar, yet belong to different classes, our method might err.
    }
  \label{fig:limitations}
\end{figure}

\noindent
\textbf{Limitations.}
\figref{fig:limitations} shows failure cases in the retrieval task.
In (a), the $5^{th}$ retrieved object is a vase, rather than a bottle. This vase is geometrically similar to a bottle, however the top differs. Most of the walks provide the geometry of a vase. There is a  lower probability of capturing this particular geometry which distinguishes it from a bottle due to the few points in that region.
In (b) we present five of the top ten objects retrieved for the leftmost nightstand. A dresser is placed at the 10th spot. These objects belong to different classes but have similar geometry.

Furthermore, our random walk generator process might be negatively impacted by strong anisotropic sampling, as occurs in 360 lidar acquisition. We have not explored this challenge and it will be an interesting direction for future work.

\section{Conclusions}

This paper introduces a novel approach for representing point clouds suited for deep learning architectures. The key idea is to represent the point cloud using multiple random walks on the shape. The randomness of the walks can be viewed as a form of data augmentation that does not require explicit manipulation of the point cloud. We utilize this representation and introduce CloudWalker, an end-to-end shape representation learning pipeline. The approach is general, yet simple, and relies on the power of randomness. We have shown our approach to be very effective for the tasks of classification and retrieval. 
An interesting avenue for future work may provide confidence measures for model’s predictions based on cross-walk correlations. Additionally, random walks in scenes would be another interesting extension of our paper. Furthermore, alternative random walk generation processes and their use in other geometrical tasks is another path to explore.

\noindent
\textbf{Acknowledgments.}
This work was partially supported by the Israel Science Foundation (ISF) 1083/18, ADRI 117632, and the European Union's Horizon 2020 research and innovation programme under the Marie Sklodowska-Curiegrant agreement No 893465.

\bibliographystyle{cag-num-names}
\bibliography{long, egbib}

\end{document}